\documentclass[letterpaper, 10 pt, conference]{ieeeconf}  

\IEEEoverridecommandlockouts                              

\makeatletter
\let\NAT@parse\undefined
\makeatother
\usepackage[colorlinks,allcolors=blue,draft=false]{hyperref}
\usepackage{amsmath} 
\usepackage{balance} 
\usepackage{amssymb}        
\usepackage{graphicx}       
\usepackage{algorithm}      
\usepackage{algpseudocode}  
\usepackage{comment}
\usepackage{siunitx}
\newif\ifanonymous
\anonymousfalse 
\usepackage{placeins}
\usepackage{standalone}
\usepackage{tikz}
\usetikzlibrary{arrows,shapes,positioning,calc,fit,backgrounds,3d,matrix}
\usepackage[capitalise]{cleveref}
\crefname{subsection}{Sec.}{Secs.}
\usepackage{xcolor}

\DeclareMathOperator*{\argmin}{arg\,min}

\title{\LARGE \bf
Spatially-Aware Adaptive Trajectory Optimization with Controller-Guided Feedback for Autonomous Racing
}

\ifanonymous
\author{The Authors\thanks{Author and funding information}}
\else
\author{Alexander Wachter$^{1}$, Alexander Willert$^{1}$, Marc-Philip Ecker$^{1,2}$ and Christian Hartl-Nesic$^{1}$%
\thanks{$^{1}$ Alexander Wachter, Alexander Willert, Marc-Philipp Ecker and Christian Hartl-Nesic are with the Automation \& Control Institute (ACIN), TU Wien, Vienna, Austria, {\tt \{wachter,willert, ecker,hartl\}@acin.tuwien.ac.at}}%
\thanks{$^{2}$Marc-Philipp Ecker is with the AIT, Austrian Institute of Technology GmbH, Vienna, Austria, {\tt ecker@ait.ac.at}}%
\thanks{$^*$ This project is funded by the FFG (FO999896399). \href{https://www.ffg.at/}{www.ffg.at}}%
}
\fi

\begin{document}

\maketitle
\thispagestyle{empty}
\pagestyle{empty}

\begin{abstract}
We present a closed-loop framework for autonomous raceline optimization that combines NURBS-based trajectory representation, CMA-ES global trajectory optimization, and controller-guided spatial feedback. Instead of treating tracking errors as transient disturbances, our method exploits them as informative signals of local track characteristics via a Kalman-inspired spatial update. This enables the construction of an adaptive, acceleration-based constraint map that iteratively refines trajectories toward near-optimal performance under spatially varying track and vehicle behavior. In simulation, our approach achieves a \SI{17.38}{\percent} lap time reduction compared to a controller parametrized with maximum static acceleration. On real hardware, tested with different tire compounds ranging from high to low friction, we obtain a \SI{7.60}{\percent} lap time improvement without explicitly parametrizing friction. This demonstrates robustness to changing grip conditions in real-world scenarios.
\end{abstract}


\section{Introduction}

High-performance autonomous racing poses a unique optimization challenge: Vehicles repeatedly traverse identical track layouts and this structured repetition can be exploited for systematic trajectory improvement. Unlike general autonomous driving on public roads where conditions vary unpredictably, racing circuits exhibit consistent spatial patterns that can be systematically learned and leveraged.
Current methods employ a decoupled two-stage paradigm: offline trajectory generation followed by online tracking control~\cite{vazquez2020optimization, ogretmen2024sampling}. Hence, conventional approaches do not exploit the repetitive nature. Recent advances focus predominantly on improving controllers to handle challenging trajectories using learning-based model-predictive control (MPC)~\cite{xue2024learning, gomes2024learning}, adaptive parameter tuning~\cite{kalaria2024adaptive}, and vehicle model adaptation~\cite{ning2023scalable, nagy2023ensemble}.

These controller-centric approaches interpret tracking errors primarily as control deficiencies and mitigate them using sophisticated strategies. As a result, they focus on local adaptations and treat each lap independently, missing the opportunity to exploit the repetitive structure of racing circuits.
Furthermore, they often require long prediction horizons to handle local disturbances as effectively as a global trajectory planner.

When a controller consistently deviates from the reference trajectory in certain track regions, it signals that the planned trajectory is the limiting factor, rather than the controller trying to track it. Instead of forcing the controller to accommodate difficult trajectories, we propose to systematically modify the trajectory globally based on feedback from controller performance.
In this way, we also shift computation from the online controller to the offline trajectory design by incorporating feedback of the controller performance into trajectory planning. This not only reduces the online computational load but also enables the controller to operate with longer horizons or additional optimization objectives.

\begin{figure}
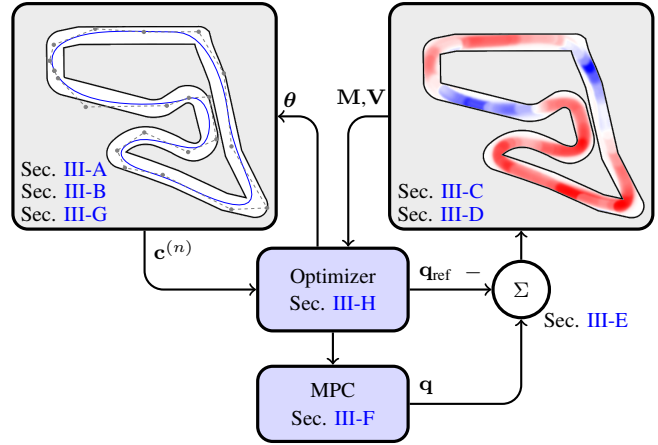

   \centering
   \includestandalone{images/overview/overview}
      \caption{Closed-loop raceline optimization framework using NURBS trajectories, CMA-ES optimization, and MPC tracking feedback.}
   \label{fig:framework_overview}
\end{figure}

In this work, we present a closed-loop raceline optimization framework that leverages controller tracking performance to improve the trajectory. Our approach constructs a spatial constraint map from the execution feedback, identifying regions where trajectory adjustments reduce the control error and lap time simultaneously, see Fig.~\ref{fig:framework_overview}. By combining NURBS-based trajectory parameterization with CMA-ES optimization, we enable continuous adaptation of both trajectory shape and timing across multiple laps, while enforcing vehicle dynamics and achieving a smooth representation of our trajectory.

Our main contributions in this work are:
\begin{itemize}
\item \textbf{Trajectory optimization based on execution feedback}: We adapt trajectory shape and timing based on controller tracking performance, moving from local controller-centric adaptation to global trajectory-centric optimization.
\item \textbf{Spatial constraint learning}: Leveraging tracking errors to learn a spatial performance map guides trajectory optimization without explicit environmental sensing.
\item \textbf{Closed-loop refinement architecture}: The integration of a NURBS-based representation with CMA-ES optimization and adaptive spatial constraints demonstrates consistent lap time improvements via trajectory adaptation.
\item \textbf{Open-source implementation}: We provide a open-source implementation with integrated GUI for ease of use. 
\end{itemize}


\vspace{-5mm}
\section{Related Work}
The field of motion planning for autonomous vehicle has progressed from basic path planning approaches to advanced, spatially-aware trajectory optimization methods, with notable developments specialized in autonomous racing.
\subsection{Motion Planning and Trajectory Optimization}
The core challenge in autonomous racing centers on the Minimum Lap Time Problem (MLTP), formulated as an optimal control problem \cite{dal2019comparison, massaro2021minimum}. Early geometric methods focused on curvature minimization \cite{brayshaw2005quasi}, while current approaches employ two-stage pipelines combining global offline optimization with local online planning \cite{betz2023tum}. Point-mass models with acceleration constraints \cite{veneri2020free, rowold2023online} offer computational speed, while single-track models with tire dynamics \cite{perantoni2014optimal, dal2018minimum} provide enhanced realism. 
\subsection{Spatially-Aware Dynamics and Adaptive Methods}
Most approaches assume spatially invariant vehicle dynamics, ignoring location-specific surface variability. The GripMap framework addresses this potential by spatially resolving dynamic constraints in Frenet coordinates, achieving \SI{5.2}{\percent} lap time improvement \cite{werner2025gripmap}. Christ et al. \cite{christ2021time} incorporate locally varying tire-road friction via friction maps, which is computationally demanding for online applications.
Model-adaptive approaches address time-varying friction without spatial resolution. Nagy et al. \cite{nagy2023ensemble} employ \emph{ensemble Gaussian processes} for multi-surface adaptation, while Kalaria et al. \cite{kalaria2024adaptive} use \emph{extreme learning machines} for online tire-slip approximation. However, these focus on temporal changes rather than position-dependent friction models leveraging repeated track traversal.
\subsection{Learning-Based Approaches and Optimization Methods}
Population-based optimization methods like CMA-ES \cite{hansen2016cma} prove effective for non-convex trajectory optimization where gradient information is uncertain.
Most of the works that solve the MLTP problem produce static trajectories without iteratively learning from execution. Recent learning-based methods have introduced significant improvements by leveraging execution feedback. \emph{Iterative learning control} (ILC) exploits repetitive lap-by-lap driving to improve feedforward inputs \cite{kapania2015path}. \emph{Learning MPC} (LMPC) extends this concept by using past lap data to refine trajectories \cite{rosolia2019learning, gomes2024learning}. Deep-kernel Gaussian processes \cite{ning2023scalable} provide enhanced vehicle dynamics modeling. Sampling-based planning \cite{ogretmen2024sampling} enables dynamic adaptation for three-dimensional race tracks.

%

\begin{figure}
    \centering
    \includegraphics[width=0.95\linewidth]{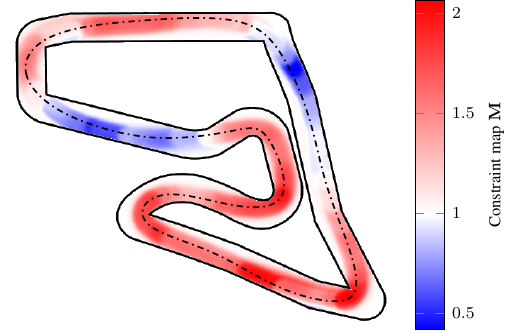}
    \caption{Constraint map visualization. Blue: reduced acceleration regions; Red: increased acceleration regions.
    }
\end{figure}

\section{Method}
We present a closed-loop learning framework for trajectory optimization in autonomous racing that integrates smooth trajectory representation, adaptive constraint modeling, and real-time control. In contrast to conventional two-step offline raceline optimization, the proposed framework formulates trajectory design as an iterative process that continuously adapts via execution feedback.

The framework comprises four main components, see Fig.~\ref{fig:framework_overview}. A \textbf{NURBS-based trajectory representation} (Section~\ref{subsec: traj rep}) restricts the search space to feasible trajectories, reducing optimization complexity. An \textbf{adaptive constraint map} (Section~\ref{subsec:adaptive constraint map}) models spatially-varying acceleration limit, which are refined via controller feedback. \textbf{Global trajectory optimization} (Section~\ref{subsec:global_traj_optimization}), implemented with CMA-ES, computes time-optimal trajectories based on the representation and constraint map.
During execution, a \textbf{tracking controller} (Section~\ref{subsec:controller}) provides feedback in terms of tracking errors. \textbf{Blame region computation} (Section~\ref{subsec:blame_region}) localizes trajectory segments associated with the tracking errors, while \textbf{error feedback} (Section~\ref{subsec:backprop}) updates the constraint map accordingly.

The design of our approach tightly couples spatial constraint adaptation and trajectory optimization, i.e., updated acceleration limits are immediately reflected in the trajectory lap time. The result is a continuous learning cycle in which execution feedback directly informs subsequent trajectory generation, leading to progressive performance improvements.

\subsection{NURBS-based Trajectory Representation}
\label{subsec: traj rep}
The foundation of our adaptive framework lies in representing the racing trajectories by Non-Uniform Rational B-Splines (NURBS), which establish a mapping from a normalized domain $u \in [0,1]$ to the raceline trajectory. This spline representation enables analytical computation of derivatives to arbitrary order, only depending on the degree of the curve. 

The spatial NURBS curve representation is mapped to a temporal trajectory with the parameterization
\begin{equation}
u(t) = \frac{t}{T}, \quad 0 \leq t \leq T~, 
\end{equation}
where $T$ denotes the lap time. This parameterization transforms the geometric path $\mathbf c(u)$ defined in the normalized domain $u \in [0,1]$ into a physical trajectory $\mathbf{q}(t) = \mathbf{c}(u(t))$ executed by the vehicle. Consequently, the vehicle's position at any time $t$ is given by evaluating the NURBS curve at the corresponding parameter value $u = t/T$.


For a NURBS curve of degree $p$ defined over $m+1$ knots $\mathbf{u} = [u_0, \ldots, u_{m}]^{\mathrm{T}}$ with $n+1$ control points $\mathbf p_i\in\mathbb R^2$, $i=0,\dotsc,n$, where $n=m-p-1$, the resulting trajectory has $C^{p-1}$ continuity and is mathematically described by
\begin{equation}
\label{eq:nurbs}
\mathbf{c}(u) = \frac{\sum_{i=0}^{n} N_{i,p}(u) w_i \mathbf{p}_i}{\sum_{i=0}^{n} N_{i,p}(u) w_i}~, \quad 0 \leq u \leq 1~,
\end{equation}
where $w_i \in \mathbb{R}^+$ represent positive weighting coefficients and $N_{i,p}(u)$ denote the B-spline basis functions of degree $p$ evaluated over the knot vector $\mathbf{u}$ given by
\begin{equation}
\label{eq: clamped vec}
\mathbf{u} = [\underbrace{0, \ldots, 0}_{p+1}, u_{p+1}, \ldots, u_{m-p-1}, \underbrace{1, \ldots, 1}_{p+1}]^{\mathrm{T}}~.
\end{equation}
Note that the knot multiplicities of $p+1$ at the boundaries allow for an analytical specification of the derivative constraints as well as the positions. In our racing context, we adopt cubic NURBS ($p=3$) to guarantee $C^2$ continuity, thereby ensuring smooth curvature transitions essential for feasibility in vehicle dynamics and control stability.

Physical derivatives of the trajectory $\mathbf{q}(t) = \mathbf{c}(t/T)$ follow from application of the chain rule, yielding
\begin{equation}
\label{eq:q_Tchainrule}
\mathbf{q}^{(k)}(t) = \frac{1}{T^k} \mathbf{c}^{(k)}\left(\frac{t}{T}\right)~, \quad k \geq 1~,
\end{equation}
where $\mathbf{c}^{(k)} = \frac{\partial^k \mathbf{c}}{\partial u^k}$ represents the $k$-th parametric derivative. This formulation not only provides computationally efficient access to velocity, acceleration, jerk, and higher-order kinematic quantities, but also allow for a dynamic limit satisfaction by utilizing the lap time $T$.

\subsection{Time-Optimal Parameterization with Spatial Adaptation}
\label{subsec. time-opt}
Using the NURBS trajectory introduced in Section~\ref{subsec: traj rep}, the velocity and acceleration along the path are expressed using \eqref{eq:q_Tchainrule} as functions of the lap time $T$. This formulation establishes the minimal feasible lap time while enforcing the vehicle's dynamic limits.

The car velocity \emph{along the path} is given by
\begin{equation}
v(t) = \frac{1}{T} \left\| \mathbf{c}^{(1)}\!\left(u\right) \right\|_2~,
\label{eq:velocity_calc}
\end{equation}
showing the scaling with $\tfrac{1}{T}$.  
By setting $v(t) = v_{\max}$ and solving for $T$, we obtain the minimum feasible lap time imposed by the velocity limit.

Similarly, the decomposition into longitudinal ($\parallel$) and lateral ($\perp$) acceleration components yields
\begin{align}
a_{\parallel}(t) &= \frac{\mathbf{c}^{(1)}(u) \cdot \mathbf{c}^{(2)}(u)}{T^2\,\|\mathbf{c}^{(1)}(u)\|_2^2}~,~~\text{and}
\label{eq:acc_long}\\
a_{\perp}(t) &= \frac{\mathbf{c}^{(1)}(u) \times \mathbf{c}^{(2)}(u)}{T^2\,\|\mathbf{c}^{(1)}(u)\|_2^2}~,
\label{eq:acc_lat}
\end{align}
where $\cdot$ and $\times$ denote the dot and cross product, respectively. 
Equations~\eqref{eq:velocity_calc}--\eqref{eq:acc_lat} 
explicitly express that velocity and acceleration scale with $\tfrac{1}{T}$ and $\tfrac{1}{T^2}$, respectively, as derived in \eqref{eq:q_Tchainrule}.
Analogous to the velocity \eqref{eq:velocity_calc}, setting $|a_{\parallel}(t)| = a_{\parallel,\max}$ and 
$|a_{\perp}(t)| = a_{\perp,\max}$ yields the corresponding lap times constrained by the longitudinal and lateral acceleration limits.

Collecting these conditions, the minimum feasible lap time under constant constraints
$v_{\max}$, $a_{\parallel,\max}$, and $a_{\perp,\max}$ is
\begin{equation}
T = \max_{u \in [0,1]} \left\{ \frac{v(u)}{v_{\max}}, \;
\sqrt{\frac{|a_{\parallel}(u)|}{a_{\parallel,\max}}}, \;
\sqrt{\frac{|a_{\perp}(u)|}{a_{\perp,\max}}} \right\}~.
\label{eq:T_const}
\end{equation}
This formulation is generalized by allowing the acceleration limits to vary along the path, i.e.,
\begin{equation}
T = \max_{u \in [0,1]} \left\{ \frac{v(u)}{v_{\max}}, \;
\sqrt{\frac{|a_{\parallel}(u)|}{a_{\parallel,\max}(\mathbf{q}(u))}}, \;
\sqrt{\frac{|a_{\perp}(u)|}{a_{\perp,\max}(\mathbf{q}(u))}} \right\}~.
\label{eq:T_spatial}
\end{equation}
The terms $a_{\parallel,\max}(\mathbf{q}(u)))$ and $a_{\perp,\max}(\mathbf{q}(u)))$ are spatially-varying functions that depend on the current position $\mathbf{q}(u)$.
These are provided by our adaptive constraint map, which is introduced in the next section.
The extension \eqref{eq:T_spatial} allows to capture location-specific phenomena directly in offline trajectory planning, such as friction variations, surface irregularities, or banking changes.
Each of these disturbances affect the lap time $T$ positively or negatively.

\subsection{Adaptive Constraint Map Framework}
\label{subsec:adaptive constraint map}
The spatially-varying acceleration limits $a_{\parallel,\max}(\mathbf{q}(u))$ and $a_{\perp,\max}(\mathbf{q}(u))$ introduced in the previous section are described by an adaptive constraint map $\mathbf{M} \in \mathbb{R}^{X \times Y}$ with $X$ and $Y$ denoting the number of grid cells horizontally and vertically. This map provides a grid-based representation of the track, where each cell value $[\mathbf M]_{x,y}=M_{x,y}$, $x=1,\dotsc,X$, $y=1,\dotsc,Y$, acts as a local scaling factor that modifies the nominal longitudinal and lateral acceleration capacities in the form
\begin{align}
	\label{eq:a_longitudinal_map}
a_{\parallel,\max}(x,y) &= M_{x,y} a_{\parallel,\text{nominal}} \\
\label{eq:a_lateral_map}
a_{\perp,\max}(x,y) &= M_{x,y}a_{\perp,\text{nominal}}~.
\end{align}
Therefore, heterogeneous track characteristics, such as friction variations, surface irregularities, or banking changes, are consistently expressed as variations in effective acceleration capability. This map is updated based on closed-loop execution data, which will be introduced in the subsequent subsections. By continuously updating the scaling factors using the trajectory-tracking feedback, the system adaptively refines local constraints, which in turn impact the computed global trajectory and its lap time, see Section~\ref{subsec. time-opt}.

\subsection{Blame Region Computation Through Derivative Analysis} \label{subsec:blame_region}
If a vehicle encounters an area of reduced friction on the track, the resulting loss of control leads to tracking errors observed later in the trajectory. This temporal and spatial separation between the cause (e.g., a low-friction area) and its impact (e.g., subsequent tracking errors) requires a systematic method to identify the origin of control degradation. Consequently, the error feedback mechanism aims to determine the causes of deviation from the planned trajectory.

\begin{figure}
	\centering
	\includegraphics{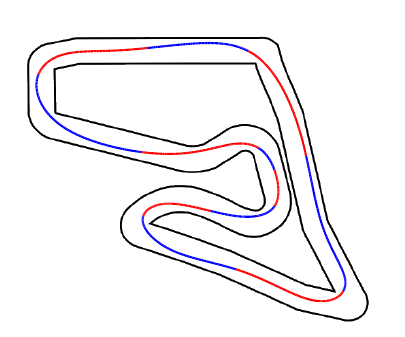}
	\caption{Track segmentation into acceleration (red) and deceleration (blue) zones for blame region identification. Since the trajectory does not contain any neutral phases, these are not represented in the figure.}
	\label{fig:acclerations_zones}
\end{figure}

Tracking errors are often linked to acceleration changes exceeding the traction limits. The longitudinal acceleration \(a_\parallel(u)\) derived from the NURBS trajectory parameterization is used to detect sign transitions using
\begin{align}
\mathcal S &= \operatorname{sign}(a_{\parallel})~, \\
\mathcal Z &= \{\, i \;\mid\; \mathcal S[i+1] \neq \mathcal S[i] \,\}~.
\end{align}
These zero crossings partition the track into zones of accelerating (positive sign), decelerating (negative sign), and neutral (zero) phases, as illustrated in Fig.~\ref{fig:acclerations_zones}.
This zoning allows errors to be traced back along the trajectory causally linked to the error.
To attribute a tracking error to its originating cause, our method associates the error with the most recent acceleration transition along the trajectory.
Specifically, the zero-crossing immediately preceding the point of minimum distance is selected. This transition index, denoted $i_{\text{transition}}$, is computed as
\begin{equation}
i_{\text{transition}}
= \max \bigl( \{\, z \in \mathcal Z \;\mid\; z < i_{\text{min}} \,\} 
\cup \{\mathcal Z[-1]\} \bigr)~.
\end{equation}
The index $i_{\text{min}}$ is defined via the closest point on the planned path $\mathbf{q}_{\text{ref}}(u)$, where the minimum distance to the actual vehicle position $\mathbf{q}$ occurs, i.e.,
\begin{equation}
	\label{eq:minimum-distance}
	\hat{e} = \min\limits_{u \in [0,1]} \|\mathbf{q} - \mathbf{q}_{\text{ref}}(u)\|_2~,
\end{equation}
which is used to quantify tracking errors.
The point $i_{\text{transition}}$ marks the boundary of the zone where the error likely originated, enabling feedback of the error to the relevant dynamic phase.

The \emph{blame region} is defined as the trajectory section between $i_{\text{transition}}$ and $i_\text{min}$. This region establishes a direct link between observed tracking errors and  
distinct trajectory characteristics, such as excessive acceleration demands, within a consistent acceleration phase. By restricting the analysis to such a region, the our method realizes targeted modifications of the trajectory where they are most effective.



\subsection{Error Feedback and Constraint Learning}
\label{subsec:backprop}

The error feedback mechanism is formulated in a general way.  
Let $ \hat{e}$ denote an general error signal obtained from the execution performance.  
This signal is modulated to promote either conservative adaptation or progressive improvement, depending on whether the observed error is below of above a given threshold $e_{\text{th}}$. Thus, the modulated error signal $e$ is defined as
\begin{equation}
e =
\begin{cases}
w^+ \hat{e}, & \text{if} \; \hat{e} < e_{\text{th}} \\
w^- \hat{e}, & \text{if} \; \hat{e} \geq e_{\text{th}}~.
\end{cases}
\end{equation}
This signal is fed back to the identified blame region, affecting all map cells $M_{x,y}$ within a radius of the relevant trajectory points.

The constraint maps \eqref{eq:a_longitudinal_map} and \eqref{eq:a_lateral_map} are updated using a method similar to a Kalman filter~\cite{simon2006optimal}, where each spatial cell $(x,y)$, $x=1,\dotsc,X$, $y=1,\dotsc,Y$, is treated as an independent state with associated uncertainty $[\mathbf V]_{x,y}=V_{x,y}$, cf.\ Fig.~\ref{fig:framework_overview}. The update equations are formulated as
\begin{align}
K_{x,y} &= \frac{V_{x,y}}{V_{x,y} + R}~, \\
M_{x,y}^+ &= M_{i,j}^- + K_{x,y}\,e~, \\
V_{x,y}^+ &= (1 - K_{x,y}) V_{x,y}^- + Q~,
\end{align}
where \(R\) is the measurement noise variance, \(Q\) is the process noise variance, and $[\mathbf K]_{x,y}=K_{x,y}$ represents the gain.
The superscripts `$-$' and `$+$' indicate values before and after the update, respectively.


\subsection{Model Predictive Control Implementation}
\label{subsec:controller}
Given a reference trajectory $\mathbf q_\mathrm{ref}(t)$, the system is controlled using a trajectory-tracking MPC. Unlike a classical MPC, which jointly optimizes path and timing, a trajectory-tracking MPC focuses solely on following the given trajectory, leveraging the embedded temporal profile \cite{yu2021model, shi2021mpc}. This decoupling simplifies the optimization process and reduces online computational effort \cite{zhang2024trajectory}, making it more suitable for real-time applications in autonomous systems.

The MPC used in this work utilizes a kinematic single-track vehicle model $\mathbf{f}(\mathbf{x}, \mathbf{u})$ with the state $\mathbf{x} = [x, y, \theta, \delta, v]^\mathrm{T}$ (comprising the position $x$ and $y$, heading $\theta$, steering angle $\delta$, and velocity $v$), and the control input $\mathbf{u} = [a, \dot{\delta}]^\mathrm{T}$ (containing the longitudinal acceleration $a$ and the steering rate $\dot\delta$) \cite{schramm2014single}. At each time step, the MPC computes control actions that minimize the deviation from the reference trajectory while respecting vehicle and actuator constraints. The resulting \emph{tracking errors} \eqref{eq:minimum-distance} provide feedback signals for adaptive trajectory adjustments in subsequent iterations.

\subsection{Trajectory Design with Closure Constraints}
\label{subsec:closure_constraints}
Our approach embeds closure constraints directly into the NURBS trajectory representation, ensuring that the optimization procedure only produces physically realizable racing trajectories. Since racing circuits are inherently closed, the trajectory must satisfy continuity conditions at the junction between the end and start of the lap.

For cubic NURBS ($p=3$), we enforce $C^2$ continuity at the closure point by analytically constraining the last three control points based on the first three. The detailed derivation of these analytical constraints is provided in the source code of our implementation. These constraints guarantee smooth transitions in position, velocity, and acceleration at the lap boundary, preventing discontinuities that would violate vehicle dynamics.
Incorporating the closure constraints reduces the optimization degrees of freedom (DoF) by three control points per spatial dimension (a total of six DoF for 2-D trajectories).


Given these constraints, the trajectory optimization operates over a reduced parameter space $\boldsymbol{\theta}$ containing only the free parameters, reading as
\begin{equation}
\label{eq:parameter_space}
\boldsymbol{\theta} = (\mathbf{p}_{\text{free}}, \mathbf{w}_{\text{free}}~, \mathbf{u}_{\text{free}})~,
\end{equation}
where $\mathbf{p}_{\text{free}} \in \mathbb{R}^{2(n-2)}$ are the control points not analytically determined by the closure constraints (i.e., $\mathbf{p}_0, \ldots, \mathbf{p}_{n-3}$), $\mathbf{w}_{\text{free}} \in \mathbb{R}^{n-2}$ are the corresponding adjustable weight parameters that control the curve locally, and
$\mathbf{u}_{\text{free}} \in \mathbb{R}^{m-2(p + 1)}$ are the interior knot positions that define the parameterization distribution along the curve. This reduced parameterization allows the optimizer to systematically explore the space of feasible trajectories while automatically satisfying continuity requirements.

\subsection{Global Trajectory Optimization}
\label{subsec:global_traj_optimization}
The NURBS representation detailed in Section~\ref{subsec: traj rep}, the adaptive local constraints detailed in Section~\ref{subsec. time-opt}, and the closure constraints in Section~\ref{subsec:closure_constraints} are solved as a global trajectory optimization problem. The optimization utilizes CMA-ES~\cite{hansen2016cma} to search the parameter space $\boldsymbol{\theta}$ from \eqref{eq:parameter_space}. The main advantages of our approach are that the lap time is implicitly solved and the optimizer can systematically only sample feasible trajectories.
The objective function for this optimization problem is formulated as
\begin{equation}
	\boldsymbol{\theta}^\ast = \argmin_{\boldsymbol{\theta}}~ 
	T(\boldsymbol{\theta}) + \lambda_{\text{dist}} \Phi_{\text{distance}} 
	+ \lambda_{\text{curv}} \Phi_{\text{curvature}}~,
\end{equation}
which combines the trajectory duration $T(\cdot)$ with geometric penalties for geometric constraint violations. 
Segments that lie outside the track boundaries are strongly penalized via $\lambda_{\text{dist}}$, which increases 
proportionally with the squared distance from the track centerline, guiding the optimizer to generate fast trajectories 
that remain within the racing limits. 
In addition, excessive curvature is penalized via $\lambda_{\text{curv}} \Phi_{\text{curvature}}$, ensuring that the local 
curvature $\kappa(u)$ does not exceed the admissible maximum $\kappa_{\max}$, which corresponds to respecting the minimal 
turning radius of the vehicle.
The curvature penalty term is chosen as
\begin{equation}
\Phi_{\text{curvature}} = \int_0^1 
 \max\bigl(0, \kappa(u) - \kappa_{\max}\bigr)^{2} \mathrm{d}u~,
\end{equation}
where $\kappa(u)$ denotes the curvature at path parameter $u$ along the trajectory.
This formulation penalizes only those segments where the curvature exceeds the 
vehicle’s maximum admissible curvature $\kappa_{\max}$.

The time scaling computation \eqref{eq:acc_long}--\eqref{eq:T_spatial} incorporates dynamic constraints using the adaptive constraint map.  
When acceleration limits reflect the learned spatial variations, the optimization process accounts for regions with reduced traction while exploiting areas with higher available grip. This coupling between trajectory optimization and constraint learning establishes a feedback mechanism in which performance of the trajectory execution influences the subsequent trajectory generation in a global manner. Additionally, an exploration component may be added, leveraging the uncertainty map $\mathbf V$ in the constraint estimates to guide the trajectory towards regions with potentially higher friction.


\section{Experimental Setup and Results}
\label{sec:experimental_setup}
\begin{figure}
	\centering
	\includegraphics[width=0.8\linewidth]{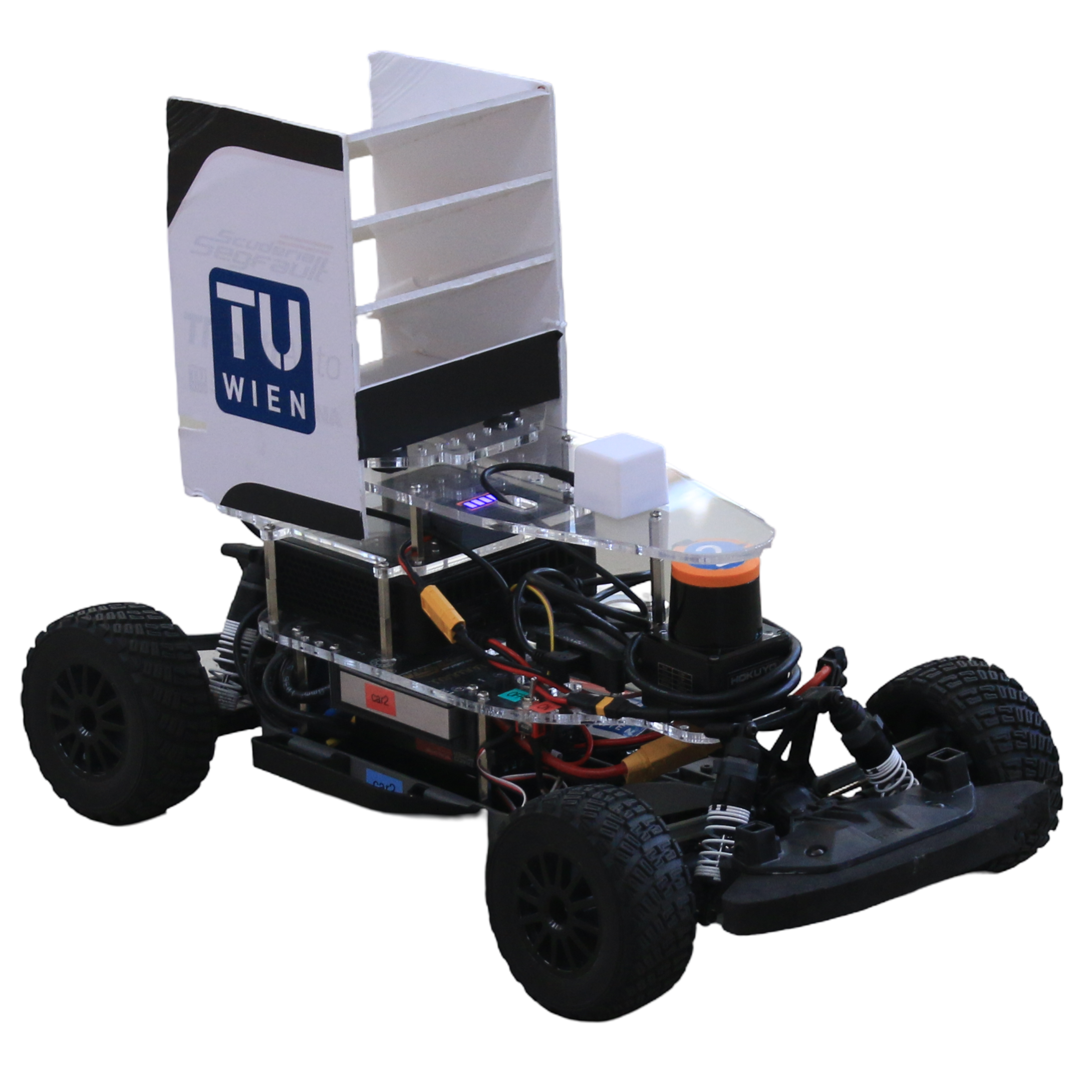}
	\caption{F1Tenth experimental platform with Asus NUC and Hokuyo LiDAR.}
	\label{fig:f1tenth_car}
\end{figure}

We evaluate our approach both in simulation and real hardware using a F1Tenth racecar \cite{o2020f1tenth}, see Fig.~\ref{fig:f1tenth_car}.
The evaluation assesses the trajectory quality, optimization performance, and the adaptation to varying surface conditions.
We compare our static as well as adaptive approach with two baselines, i.e.,
(i) a minimum-curvature trajectory (\emph{Min Curvature}) \cite{heilmeier2020minimum}, 
(ii) a raceline optimized using an approximate homeomorphism (\emph{Homeomorphism}) \cite{klapalek2021car}.

First, all baseline approaches are evaluated in simulation on a variety of static racetracks~\cite{charles2025advancing} to assess the quality of the generated trajectories and the trajectory representation.
Second, we measure lap times with and without adaptive approach.
Third, on a racetrack with low-friction regions we run a detailed analysis to quantify how quickly our approach is able to adapt.
Fourth, we conduct real-world experiments on a track with varying surface properties. 
In this hardware experiment, global friction is varied using different wheel sets, enabling us to evaluate how effectively our method adapts to varying friction limits along the trajectory.

\subsection{Performance Comparison}

We evaluate the impact of our method on lap performance across four distinct track configurations with varying complexity and length. Table~\ref{tab:traj_times_vs_baseline} presents the lap times for both the baseline optimizations and our approach with and without adaptation.
\begin{table}
	\centering
	\caption{Lap times in \si{\second} for different trajectories and racetracks.}
	\label{tab:traj_times_vs_baseline}
	\setlength{\tabcolsep}{4pt}
	\begin{tabular}{c|cccc}
		\hline
		Map & Ours    & Ours  & Min Curvature & Homeomorphism \\
        & Static & Adaptive & Static & Static \\
        \hline
		F1Aut     & 20.02 & \textbf{16.54} & 22.12 & 21.83 \\
		Wall1      & 16.82 & \textbf{15.71} & 17.32 & 16.92 \\
		Levine     & 11.08 & \textbf{10.42} & 11.39 & 12.18 \\
		Operngasse &  7.49 &  \textbf{6.24} &  8.37 & 7.58  \\
		\hline
	\end{tabular}
\end{table}
The results demonstrate consistent performance improvements across all tested circuits. Our adaptation method improves the lap time from \SI{16.82}{\second} to \SI{15.71}{\second} on the Wall1, which corresponds to a relative improvement of \SI{6.60}{\percent}. The most significant improvements are observed on the long F1Aut track with high-speed corners, where our approach reduces the lap time from \SI{20.02}{\second} to \SI{16.54}{\second}, corresponding to an improvement of \SI{17.38}{\percent}. The smaller Operngasse and Levine tracks show improvements of \SI{16.69}{\percent} and \SI{5.96}{\percent} respectively, indicating that the benefits scale effectively across different track characteristics. 
Compared to the \emph{Min Curvature} method, which yields lap times of \SI{22.12}{\second} (F1Aut), \SI{17.32}{\second} (Wall1), \SI{11.39}{\second} (Levine), \SI{8.37}{\second} (Operngasse), respectively, our approach consistently achieves faster laps. In contrast, the lap times corresponding to the \emph{Homeomorphsim} approach exhibit up to \SI{9.93}{\percent} performance loss against our static approach. The performance gains of our approach listed in Table~\ref{tab:traj_times_vs_baseline} are achieved while maintaining constant tracking error throughout the trajectory, ensuring that the improved lap times do not compromise vehicle control and safety.
\vspace{-1mm}
\subsection{Impact of Adaptive Constraint Map}
\begin{figure}
	\centering
	\includegraphics{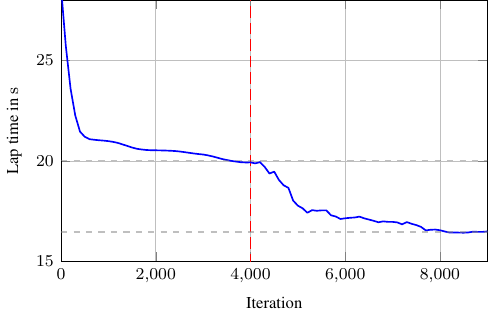}
	\caption{Optimizer convergence on F1Aut track. Red line: Error feedback activation. Convergence is achieved within 8~laps, with 500~iterations/lap.}
	\label{fig:Trajectory duration over iteration}
\end{figure}

The impact of the adaptive constraint map described in Sections~\ref{subsec:adaptive constraint map} to~\ref{subsec:backprop} is shown in Fig.~\ref{fig:Trajectory duration over iteration}, which depicts the convergence behavior in simulation for the F1Aut track.
The first $4000$ optimizer iterations are completed without providing error feedback to our optimizer. The car converges to a lap time of \SI{20.02}{\second}. As soon we enable the feedback indicated by the vertical dashed redline, the optimizer converges after $8$ laps (roughly 500$\,$iterations/lap) to a lap time of \SI{16.54}{\second}, a relative improvement of \SI{17.38}{\percent}. 

\subsection{Convergence Analysis with Local Friction Variations}
\begin{figure*}
	\centering
	\includegraphics{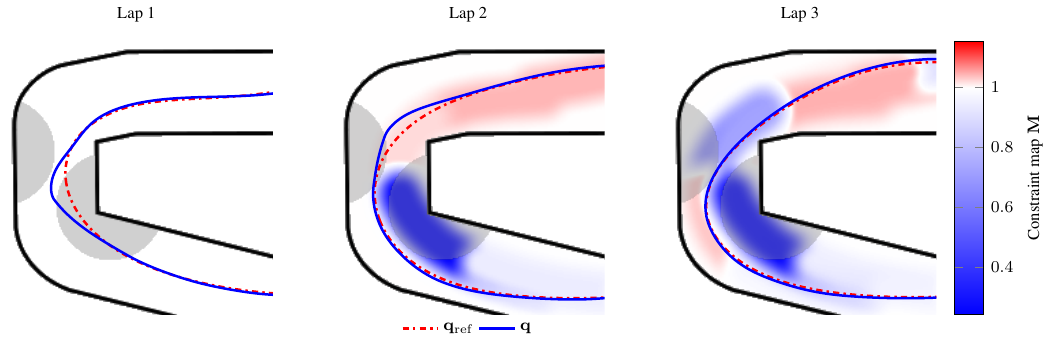}
	\caption{llustration of the two artificially introduced low-friction regions (highlighted in gray) and the corresponding evolution of the optimized trajectory across three laps, showing the nominal planned trajectory (red-dotted line), the executed trajectory (blue line), and updated constraint map (shaded blue and red regions).}
	\label{fig:Local friction}
\end{figure*}

To evaluate the optimizer's responsiveness to localized surface variations, we introduce two regions with \SI{80.0}{\percent} less friction compared to the nominal track surface. Fig.~\ref{fig:Local friction} illustrates these regions as well as the evolution of the optimized trajectory over three laps, in which the spatial convergence occurs.
Initially unaware of friction variations, the vehicle traverses the first low-friction region using the nominal trajectory in lap $1$. After feeding back the tracking error between the red-dotted reference trajectory $\mathbf q_\mathrm{ref}$ and the executed blue trajectory $\mathbf q$, the blame region (shown in blue) is updated. The optimizer then adapts both the trajectory shape and its timing, enabling the vehicle to successfully navigate the second previously unknown low-friction area during lap~2, where the same procedure is repeated. By lap $3$, the optimizer converges to an optimal path that effectively navigates the reduced-grip corner, indicated by the blue line. This demonstrates quick adaptation to localized surface variations of our method.

\subsection{Real-World Adaptive Experiments}
We validated our approach using the real-time system depicted in Fig.~\ref{fig:f1tenth_car} and we documented the experiments as video 
\ifanonymous
in the supplementary material.
\else
at \href{-}{(will be done after until the video-submission deadline)}.
\fi  
To demonstrate the impact of the proposed method, we tested global friction variations by employing different tire configurations on the vehicle. Three scenarios were evaluated: a low-friction configuration with four off-road tires (\emph{Low}), a medium-friction configuration with slick front tires and off-road rear tires (\emph{Medium}), and a high-friction configuration with four slick tires (\emph{High}).
 
\begin{figure}
	\centering
	\includegraphics{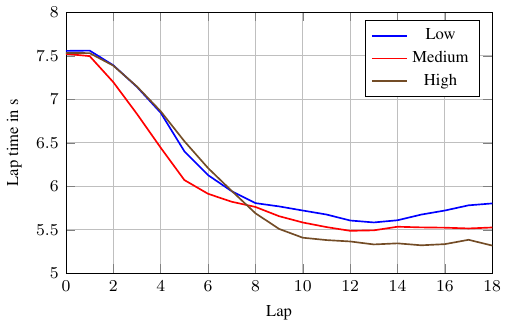}
	\caption{Lap times for three tire configurations (Low/Medium/High friction). Error feedback activated at lap 0.}
	\label{fig:global_friction lab times}
\end{figure}

\begin{figure*}
	\centering
	\includegraphics[scale=0.97]{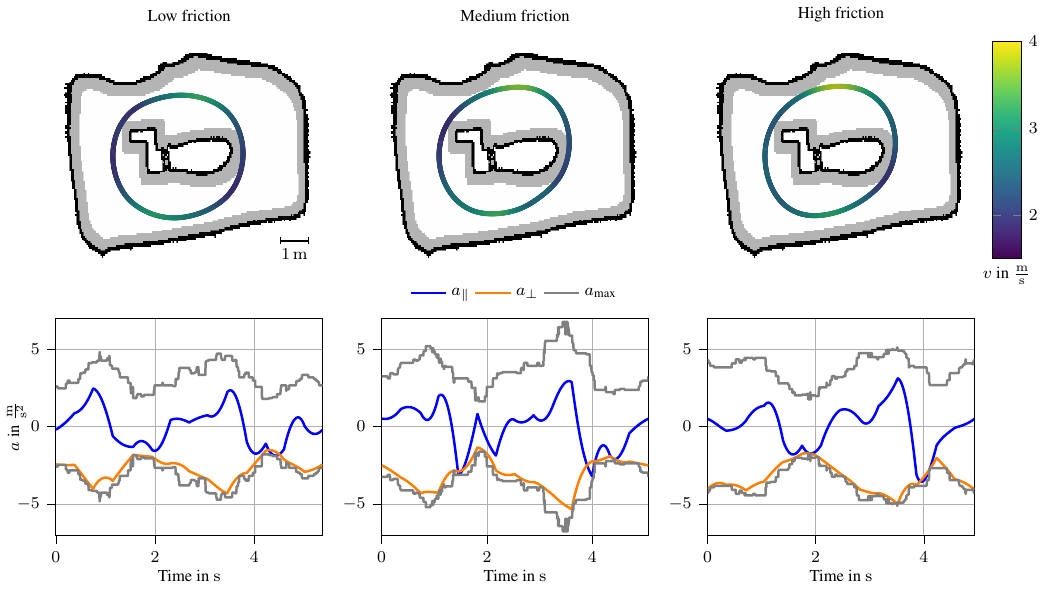}
	\caption{Optimized trajectories under different global friction conditions showing the lateral acceleration $a_{\perp}$, the longitudinal acceleration $a_{\parallel}$, and the acceleration constraints $a_\mathrm{max}$ from the constraint map. Higher-friction setups produce trajectories with increased average velocity and slightly longer path lengths, while lower-friction setups result in reduced speeds and more circular trajectories. 
	}
	\label{fig:global_friciton_lap}
\end{figure*}

The baseline performance without friction parameterization yielded a conservative lap time of \SI{7.53}{\second} on a circular track with varying width, as shown in Fig.~\ref{fig:global_friction lab times}.
Upon activation of the error-feedback mechanism, consistent performance improvements were observed across all tire configurations. In order to achieve convergence, the error feedback threshold was set based on the position measurement accuracy. Therefore, a balance between position measurement noise and maximum admissible deviation from the reference trajectory was chosen for $e_{\text{th}}$, with convergence being achieved after approximately $10$ laps. The system achieved a minimum lap time of \SI{5.29}{\second} under high-friction conditions, while the medium and low friction configurations converged to lap times of \SI{5.56}{\second} and \SI{5.73}{\second}, respectively. This corresponds to an adequate adaptation to the different friction conditions, resulting in a \SI{7.60}{\percent} lap time improvement without explicit friction parametrization or environment sensing.
The resulting trajectory shapes after convergence are illustrated in Fig.~\ref{fig:global_friciton_lap}.

Due to the circuit layout, vehicle performance was primarily constrained by lateral acceleration $a_{\perp}$. Fig.~\ref{fig:global_friciton_lap} also shows the exploration of the local acceleration limits while maximizing their utilization. For higher friction conditions, the algorithm generates trajectories with increased average velocity and slightly greater path length, whereas lower friction configurations result in reduced speed and a more circular trajectory shape. Note that when acceleration limits are set to maximum or mean acceleration values, the controller fails to accurately follow the generated trajectory, resulting in performance degradation or, in worst-case scenarios, vehicle instability resulting in a crash.
Notably, the system does not require manual tuning of the constraint parameters. Instead, it automatically discovers the feasible limits for each wheel configuration.

\vspace{-5mm}
\section{Conclusion}\label{sec:conclusion}
The closed-loop raceline optimization framework proposed in this work systematically leverages execution feedback to global trajectory improvements for autonomous racing applications, shifting from controller-centric adaptation to trajectory-centric optimization. Our proposed approach integrates NURBS trajectory parameterization with CMA-ES optimization and learning of adaptive spatial constraints. This allows continuous refinement of both trajectory shape and timing using the MPC tracking error as feedback. The spatial-constraint learning mechanism translates tracking errors into location-specific constraint maps without explicit environmental sensing, while performing causal attribution of errors to specific trajectory segments using a so-called \emph{blame region} computation.

Experimental validation demonstrates consistent lap time improvements ranging from \SI{5.96}{\percent} to \SI{17.38}{\percent} across diverse track configurations in simulation and real-world F1Tenth experiments. We achieve convergence after approximately 10 laps and performance improvements across varying friction conditions. The proposed method shows significant performance enhancement compared to static trajectory approaches, highlighting its potential in applications for autonomous vehicle where structured repetition enables systematic learning.

Future work will investigate controller dependency by systematic evaluation with different underlying control architectures, extend beyond positional error metrics to incorporate temporal dependencies, and replace the Kalman-inspired update mechanism with learning-based approaches that automatically identify origin regions of error.

\ifanonymous
\else
\section*{Acknowledgement}
Acknowledgement and funding information.
(will be done after until the submission deadline).
\fi  



\bibliographystyle{class/IEEEtran}
\bibliography{class/IEEEabrv,class/reference}
\vspace{19mm}
\pagebreak

\end{document}